\documentclass{article}
\usepackage[table,xcdraw]{xcolor}

\usepackage{PRIMEarxiv}

\usepackage[utf8]{inputenc} %
\usepackage[T1]{fontenc}    %
\usepackage{hyperref}       %
\usepackage{url}            %
\usepackage{booktabs}       %
\usepackage{amsfonts}       %
\usepackage{nicefrac}       %
\usepackage{microtype}      %
\usepackage{lipsum}
\usepackage{fancyhdr}       %
\usepackage{graphicx}       %
\graphicspath{{media/}}     %
\usepackage[utf8]{inputenc} %
\usepackage[T1]{fontenc}    %
\usepackage{hyperref}       %
\usepackage{url}            %
\usepackage{booktabs}       %
\usepackage{amsfonts}       %
\usepackage{nicefrac}       %
\usepackage{microtype}      %
\usepackage[most]{tcolorbox} 
\usepackage{tabularx}   %
\usepackage{booktabs}   %
\usepackage{multirow}
\usepackage{array} 
\usepackage{pdfpages}
\usepackage{float}
\usepackage{natbib}

\definecolor{bestblue}{RGB}{180, 210, 245} 
\definecolor{secondblue}{RGB}{224, 238, 255}

\title{On Path to Multimodal Historical Reasoning: HistBench and HistAgent

}

\begin{document}

\renewcommand{\author}{}
\maketitle

\vspace{-9.3em}
\section*{Organizing Team}
\vspace{-0.5em}
\noindent 
Jiahao Qiu$^{*1}$, Fulian Xiao$^{*2}$, Yimin Wang$^{*3}$, Yuchen Mao$^{*4}$, Yijia Chen$^{*5}$, 
Xinzhe Juan$^3$, Shu Zhang$^{11}$, Siran Wang$^2$, Xuan Qi$^6$, Tongcheng Zhang$^4$, Zixin Yao$^7$, Jiacheng Guo$^1$, Yifu Lu$^1$, Charles Argon$^8$, Jundi Cui$^2$, Daixin Chen$^5$, Junran Zhou$^2$, Shuyao Zhou$^1$, Zhanpeng Zhou$^4$, Ling Yang$^1$, Shilong Liu$^1$, Hongru Wang$^9$, Kaixuan Huang$^1$,\\ Xun Jiang$^{10,11}$, Xi Gao$^{\dagger2}$, Mengdi Wang$^{\dagger1}$

\vspace{-0.2em}
\noindent
\textbf{$^1$AI Lab, Princeton University} \quad
\textbf{$^2$Department of History, Fudan University} \quad
\textbf{$^3$University of Michigan} \\
\textbf{$^4$Shanghai Jiao Tong University} \quad
\textbf{$^5$School of Philosophy, Fudan University} \quad
\textbf{$^6$IIIS, Tsinghua University} \\
\textbf{$^7$Department of Philosophy, Columbia University} \quad
\textbf{$^8$Department of History, Princeton University} \\
\textbf{$^9$The Chinese University of Hong Kong} \quad
\textbf{$^{10}$Tianqiao and Chrissy Chen Institute} \quad
\textbf{$^{11}$Theta Health Inc.}

\vspace{-0.8em}
\section*{Dataset Contributors}
\vspace{-1em}
\noindent
Yuming Cao, Yue Chen, Yunfei Chen, Zhengyi Chen, Ruowei Dai, Mengqiu Deng, Jiye Fu, Yunting Gu, Zijie Guan, Zirui Huang, Xiaoyan Ji, Yumeng Jiang, Delong Kong, Haolong Li, Jiaqi Li, Ruipeng Li, Tianze Li, Zhuoran Li, Haixia Lian, Mengyue Lin, Xudong Liu, Jiayi Lu, Jinghan Lu, Wanyu Luo, Ziyue Luo, Zihao Pu, Zhi Qiao, Ruihuan Ren, Liang Wan, Ruixiang Wang, Tianhui Wang, Yang Wang, Zeyu Wang, Zihua Wang, Yujia Wu, Zhaoyi Wu, Hao Xin, Weiao Xing, Ruojun Xiong, Weijie Xu, Yao Shu, Yao Xiao, Xiaorui Yang, Yuchen Yang, Nan Yi, Jiadong Yu, Yangyuxuan Yu, Huiting Zeng, Danni Zhang, Yunjie Zhang, Zhaoyu Zhang, Zhiheng Zhang, Xiaofeng Zheng, Peirong Zhou, Linyan Zhong, Xiaoyin Zong, Ying Zhao, Zhenxin Chen, Lin Ding, Xiaoyu Gao, Bingbing Gong, Yichao Li, Yang Liao, Guang Ma, Tianyuan Ma, Xinrui Sun, Tianyi Wang, Han Xia, Ruobing Xian, Gen Ye, Tengfei Yu, Wentao Zhang, Yuxi Wang

\vspace{2em}

\begingroup
\let\thefootnote\relax
\footnotetext{$^*$ Equal contribution.}
\footnotetext{$^\dagger$ Correspondence to: \texttt{gaoxi@fudan.edu.cn}, \texttt{mengdiw@princeton.edu}.}
\endgroup

\begin{abstract}
Recent advances in large language models (LLMs) have led to remarkable progress across various domains, yet their capabilities in the humanities, particularly history, remain underexplored. Historical reasoning poses unique challenges for AI, involving multimodal source interpretation, temporal inference, and cross-linguistic analysis. While general-purpose agents perform well on many existing benchmarks, they lack the domain-specific expertise required to engage with historical materials and historical questions. To address this gap, we introduce HistBench, a new benchmark of 414 high-quality questions designed to evaluate AI's capacity for historical reasoning and authored by more than 40 expert contributors. The tasks span a wide range of historical problems—from factual retrieval based on primary sources to interpretive analysis of manuscripts and images, to interdisciplinary challenges involving archaeology, linguistics, or cultural history. Furthermore, the benchmark dataset spans 29 ancient and modern languages and covers a wide range of historical periods and world regions. Finding the poor performance of LLMs and other agents on HistBench, we further present HistAgent, a history-specific agent equipped with carefully designed tools for OCR, translation, archival search, and image understanding in History. On HistBench, HistAgent based on GPT-4o achieves an accuracy of 27.54\% pass@1 and 36.47\% pass@2, significantly outperforming LLMs with online search and generalist agents, including GPT-4o (18.60\% ), DeepSeek-R1(14.49\% ), Grok 3(17.63\%) and Open Deep Research-smolagents(20.29\% pass@1 and 25.12\% pass@2). These results highlight the limitations of existing LLMs and generalist agents and demonstrate the advantages of HistAgent for historical reasoning. Notably, HistAgent also achieves 60.00\% pass@1 accuracy on GAIA, showing that domain-specific customization doesn't hinder HistAgent's competitive performance on real-world general tasks. 
\\
\\
\textbf{Code: }\url{https://github.com/CharlesQ9/HistAgent}\\
\textbf{Dataset: }\url{https://huggingface.co/datasets/jiahaoq/HistBench}
\end{abstract}

\begin{figure}[H]
  \centering
  \includegraphics[width=0.9\textwidth]{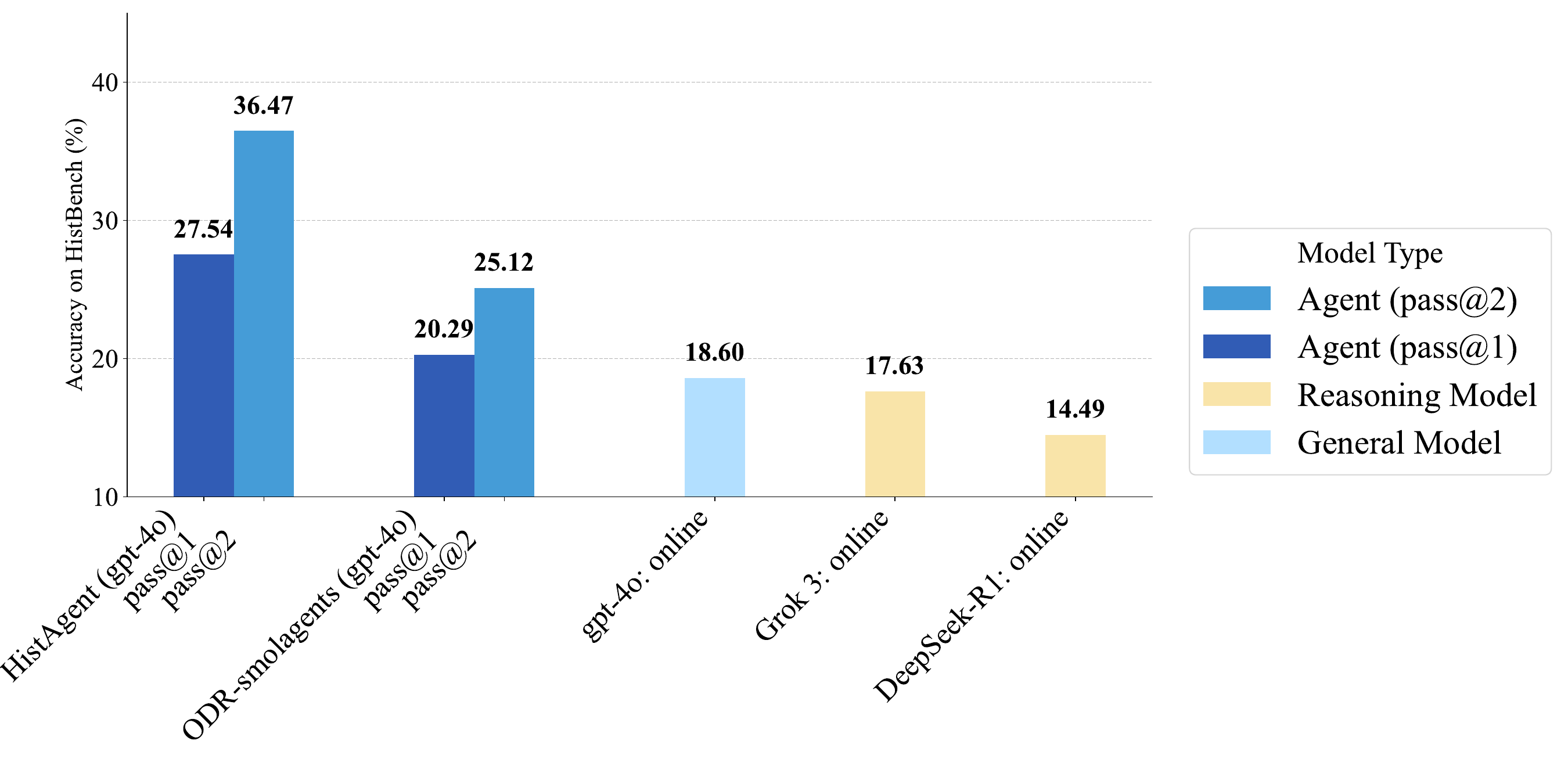}
  \caption{Performance of LLMs and Agents on HistBench.}
  \label{fig:performance on HistBench}
\end{figure}

\section{Introduction}

Recent advances in large language models (LLMs) have led to AI agents capable of impressive performance across a range of complex tasks. Although significant progress has been made in general-purpose and AI4Science agents, such as OpenAI’s Deep Research \citep{openai2025introducing} and agents for chemistry and biology~\citep{huang2024crispr,bran2023chemcrow}, the humanities remain critically underexplored. This imbalance risks narrowing both the scope of artificial intelligence and our understanding of intelligence itself. If AI is to become truly general, it must also engage with the complex and underexplored challenges posed by the humanities.

Among the humanities, history holds a uniquely central role. It addresses fundamental questions of human identity, continuity, and transformation, while exemplifying the interpretive complexity of humanistic scholarship. Historical reasoning requires navigating incomplete, heterogeneous sources—ranging from manuscripts and inscriptions to maps and visual records—and demands cross-linguistic, cross-modal, and cross-cultural interpretation. Unlike factual recall, it involves temporal reasoning, contextualization, and the reconciliation of conflicting narratives. As both epistemically rich and methodologically demanding, history encapsulates many of the core challenges that characterize reasoning in the humanities. In addition to its disciplinary significance, history is also the most computationally tractable field among the humanities. It generates and consumes large-scale textual and visual data, far exceeding most other humanities domains in volume, structure, and annotation potential. Moreover, historical research is inherently global and multilingual, shaped by regionally specific contexts. These features make history an ideal testbed for evaluating AI systems under real-world challenges such as cross-lingual retrieval, multi-modal reasoning, and cultural alignment. For these reasons, history is not only methodologically appropriate but also strategically positioned as the first point of engagement for AI-humanities research.

Despite the proliferation of AI benchmarks, there remains no rigorous evaluation suite specifically designed to assess historical reasoning. Most existing benchmarks either focus on general capabilities or are tailored to domains like science and mathematics. For example, the GAIA benchmark evaluates real-world tasks across broad domains, but does not target any single discipline in depth~\citep{mialon2023gaia}. Humanity’s Last Exam (HLE), a large-scale human-written test covering over 2,500 questions, includes only 56 history-related problems~\citep{phan2025humanity}—far too few to evaluate the interpretive, evidentiary, and temporal reasoning skills central to historical reasoning. Meanwhile, discipline-specific benchmarks such as PhyBench~\citep{qiu2025phybench} demonstrate the importance of domain-adapted evaluation in fields like physics, but no equivalent exists for history. As a result, there is currently no benchmark that captures the unique methodological demands of historical research and historical reasoning, leaving a critical gap in evaluating LLM and agent performance in this foundational area of the humanities.

To address this gap, we introduce HistBench, the first comprehensive benchmark dedicated to evaluating historical reasoning in AI systems. The dataset contains 414 high-quality and carefully-reviewed questions, contributed by history professors, PhD students, and research assistants. Each question simulates a realistic research task and is annotated with difficulty level, question type, and reasoning dimension. The tasks span a wide range of historical problems—from factual retrieval based on primary sources to interpretive analysis of manuscripts and images, to interdisciplinary challenges involving archaeology, linguistics, or cultural history. HistBench questions are categorized into three difficulty levels—Level 1 (Basic), Level 2 (Intermediate), and Level 3 (Challenging)—based on six structured criteria: rarity of source knowledge, linguistic complexity, format heterogeneity, perceptual accessibility, interdisciplinary scope, and reasoning depth. The benchmark covers 29 languages and all major world regions, and all questions are carefully designed to resist resolution through retrieval-based answering or shallow prompting. By combining topical breadth with rigorous, source-grounded construction, HistBench enables domain-specific evaluation of LLMs under expert-defined historical reasoning constraints.

With the introduction of HistBench, we can now systematically evaluate how well LLMs and generalist agents perform on historical tasks. Our findings reveal a consistent gap: current large language models and agents lack the capabilities required for historical reasoning. Unlike other tasks, historical research involves interpreting fragmented, multimodal sources across languages, formats, and time periods. However, today’s models and agents fall short in meeting the demand for historical reasoning. Their OCR tools fail on handwritten manuscripts, damaged archives, and early print formats; their translation capabilities are limited to high-resource modern languages, with poor support for classical and minority languages; and their visual understanding of historical materials—such as annotated maps, marginalia, or iconography—is minimal. These limitations are not only technical but methodological: historical analysis demands multi-step reasoning, cross-source synthesis, and epistemic grounding that current general-purpose systems do not support. This performance gap underscores the need for dedicated agents equipped with domain-specific tools and workflows aligned with historical practice.

To overcome the limitations of large language models and generalist agents, we introduce HistAgent, a domain-specialized AI agent tailored for historical reasoning. Built on top of an advanced LLM, HistAgent integrates a set of modular tools designed to address the epistemic and technical challenges unique to historical inquiry. These include OCR modules for handwritten manuscripts (Transkribus, Asian-script OCR), multilingual translators with provenance preservation, reverse image search for historical visuals, and peer-reviewed literature retrieval mimicking history researchers. HistAgent spans multiple modalities—text, image, audio, manuscript, video—and applies source-aware workflows for extraction, parsing, and reasoning.

On HistBench, HistAgent based on GPT-4o achieves an accuracy of 27.54\% pass@1 and 36.47\% pass@2, significantly outperforming LLMs with online search and generalist agents, including GPT-4o (18.60\% ), DeepSeek-R1 (14.49\% ), Grok 3(17.63\%) and open Deep Research by smolagents(20.29\% pass@1 and 25.12\% pass@2). These results highlight the limitations of existing LLMs and generalist agents and demonstrate the advantages of HistAgent for historical reasoning. Notably, HistAgent also achieves 60.00\% pass@1 accuracy on the GAIA benchmark, showing that domain-specific customization doesn't hinder HistAgent's competitive performance on real-world general tasks.

In summary, this paper makes the following contributions:
\begin{itemize}
    \item \textbf{HistBench}: We present HistBench, the first large-scale and comprehensive benchmark involving 414 high-quality and carefully-reviewed questions for evaluating historical reasoning in AI systems. 
    
    \item \textbf{HistAgent}: We present HistAgent, an agent specialized for the history domain. HistAgent integrates multi-modal perception, web browsing, and a suite of modular tools—including OCR for manuscripts, image provenance retrieval, multilingual translation, literature parsing, and document analysis—into a cohesive framework for historical research. It is designed to mirror the interpretive workflow of human historians, enabling document-level reasoning, multimodal synthesis, and context-aware analysis across languages and source formats. By combining general LLM capabilities with domain-specific customization, HistAgent overcomes key limitations of existing LLMs and generalist agents in handling historical tasks while maintaining competitive performance on general tasks.
    
    \item \textbf{Comprehensive Empirical Validation}: We provide an extensive evaluation of HistAgent on both domain-specific and general benchmarks. HistAgent significantly outperforms base LLMs with online search on HistBench and the history subset of Humanity's Last Exam, demonstrating superior historical reasoning. Furthermore, HistAgent achieves high performance on real-world general-purpose benchmarks like GAIA, highlighting that a history-focused agent can maintain broad capabilities and excel in complex, multi-step reasoning tasks beyond its core domain.

\end{itemize}

By introducing HistBench and HistAgent, this work lays a concrete foundation for the path toward multi-modal historical reasoning in AI. Our benchmark captures the complexity of historical inquiry across languages, media, and reasoning styles, while our agent demonstrates how agents can begin to emulate the interpretive workflows of human historians. We hope this effort not only advances AI for history, but also inspires broader exploration of LLMs and domain-specific agents across other underrepresented areas of human knowledge and marks a broader step toward AI for the Humanities.

\section{Related Works}
\label{app:related_works}

\subsection{Generalist Agent}

Generalist agent is for solving general real-world tasks such as GAIA\citep{mialon2023gaia}. OMNE~\citep{jiang2024long} introduces a column-structured long-term memory that agents update during inference to refine policies without retraining. AutoAgent~\citep{tang2025autoagent} compiles natural-language workflow descriptions into executable multi-agent pipelines. OWL~\citep{owl2025} adds structured orchestration using a CAMEL-based Orchestrator–Worker pattern, where the orchestrator delegates subtasks to specialists via explicit transfer actions. For information-dense tasks, OpenAI Deep Research~\citep{openai2025introducing} combines web browsing, document parsing, and grounded synthesis to produce cited reports, while open Deep Research by Smolagents \citep{smolagents} reproduces the same workflow in the open‑source domain using a Python‑driven CodeAgent that reduces communication overhead. They jointly define today’s design space for scalable generalist multi‑agent solutions.

\subsection{Domain-specific Agent}
Domain-specific agents address the limitations of general-purpose LLMs in tasks requiring deep expertise. A primary approach to imbue such specificity is parametric adaptation through fine-tuning on domain-centric corpora, enabling models to learn specialized terminology, data patterns, and reasoning pathways. Notable examples include CRISPR-GPT~\citep{huang2024crispr}, an agent designed to automate and enhance the design of CRISPR-based gene-editing experiments by leveraging domain knowledge and external tools, BrainGPT~\citep{li2025braingpt}, a model clinically visual instruction-tuned for 3D brain CT radiology report generation , and DeepSeekMath~\citep{shao2024deepseekmathpushinglimitsmathematical}, which demonstrates advanced mathematical reasoning after extensive training on mathematical texts. These methodologies are crucial for creating agents that are not only knowledgeable but also more reliable and effective within their specific operational contexts.

Beyond foundational knowledge embedding, the efficacy of domain-specific agents is significantly amplified by their ability to interact with and act upon their environment through tool integration. ChemCrow~\citep{bran2023chemcrow}, for instance, leverages a suite of 18 expert-designed chemistry tools to perform complex tasks in organic synthesis, drug discovery, and materials design, showcasing enhanced performance in chemistry-related problem-solving. Other advanced approaches include Case-Based Reasoning (CBR), enabling agents such as DS-Agent~\citep{guo2024dsagent} to learn from past expert solutions in fields like automated data science. WarAgent~\citep{hua2023war} exemplifies domain-specific simulation of macro-scale historical conflicts using LLM-based multi-agent systems, allowing country agents to reenact international relations dynamics and explore war triggers and outcomes. Similarly, BattleAgent~\citep{lin2024battleagent} extends this line of work by incorporating multimodal inputs and fine-grained modeling of individual soldiers in historical combat, demonstrating how agent-based frameworks can reconstruct both strategic decisions and personal experiences in complex environments. These approaches collectively aim to produce agents that are not only competent in their domain but also robust and trustworthy in real-world settings.

\subsection{Domain-specific Benchmarks}

The evaluation of large language models (LLMs) is moving from broad assessments toward domain-specific benchmarks (DSBs). General-purpose tests often miss the detailed strengths and weaknesses of LLMs when applied in real-world settings. Common issues with standard benchmarks include limited coverage of field knowledge, poor alignment with practical tasks, vulnerability to data contamination that encourages memorization, and simple question formats that do not test multi-step or advanced reasoning \citep{Wang2023SciBench, alaa2025medical, Ahn2024RVBench, bodensohn2025unveiling}. Questions about data quality and contamination have led to efforts such as HLE, which invests in carefully curated, contamination-resistant items \citep{phan2025humanity}. As a result, the number of DSBs has grown rapidly. These new tests help guide model improvements, set realistic expectations for deployment, and pinpoint specific model capabilities in different areas \citep{Guha2023LegalBench, pipitone2024legalbench}.

This trend towards specialization is evident across numerous domains. In software engineering, benchmarks like SWE-Bench \citep{Jimenez2023SWEBench} evaluate LLMs on resolving real-world GitHub issues, moving beyond simpler code generation tasks assessed by benchmarks like HumanEval or MBPP \citep{Chen2021HumanEval, Austin2021MBPP, Jimenez2023SWEBench, phan2025humanity}. The medical domain utilizes benchmarks such as PubMedQA \citep{Jin2019PubMedQA} and the MultiMedQA suite \citep{Singhal2023MultiMedQA} to assess medical knowledge and reasoning, although ongoing research seeks to improve alignment with clinical practice and address benchmark saturation \citep{Singhal2023MultiMedQA, alaa2025medical}. Legal AI evaluation employs benchmarks like LegalBench \citep{Guha2023LegalBench} for diverse legal reasoning tasks  and specialized versions like LegalBench-RAG \citep{pipitone2024legalbench} focusing on retrieval-augmented generation crucial for fact-intensive legal work. For scientific and mathematical reasoning, SciBench \citep{Wang2023SciBench} presents collegiate-level problems , RV-Bench \citep{Ahn2024RVBench} targets genuine mathematical understanding , and PHYBench~\cite{qiu2025phybench} specifically assesses complex physics reasoning using problems from global exams and competitions. Finance has seen the development of benchmarks like FinanceQA \citep{mateega2025financeqa}, tailored to evaluate performance on tasks mirroring real-world financial analysis. Collectively, these domain-specific evaluations are crucial for probing deeper LLM capabilities and fostering the development of models suitable for specialized professional deployment.

\subsection{History Benchmarks} 
In the domain of history, based on a subset of the Seshat Global History Databank, HiST-LLM evaluates the possession of expert historical knowledge of Seven Models.~\citep{hauser2024large}
However,there is a discrepancy between historical knowledge and historical research.HLE can be regarded as a combination of many domain-specific benchmarks. In terms of the historical questions in HLE, although the capabilities of LLMs in historical research is indeed evaluated (as opposed to the benchmarks mainly focused on historical knowledge), the total number of questions is only 56.~\citep{phan2025humanity} For existing benchmarks, there are limitations in both the scope of the questions and the disciplinary characteristics.

We need to detail several limitations in these benchmarks. The creators of Gaia~\citep{mialon2023gaia} indicate that Gaia has three limitations—Missing evaluations: Shortage of evaluation of trace leading to the answer; On the cost of designing unambiguous questions: Shortage of ambiguous questions that may appear in the daily usage scenario; Lack of linguistic and cultural diversity: Shortage of questions in languages other than standard English.The creators of HiST-LLM also indicate the limitations of the database they use (the subset of the Seshat Gloval History Databank). First, their data are mostly from sources in English;Second, the expertise and background of the research assistants may influence the definition of variables. Third, the benchmark is only the reflection of the current recognition. In addition to these limitations, it is necessary to repeat that HiST-LLM is a benchmark designed for the evaluation of the possession of historical knowledge instead of the capabilities of historical research such as literature retrieval, historical source retrieval, historical analysis, historical source processing, and interdisciplinary.

Compared to the comprehensive benchmark like HLE, an independent domain-specific benchmark can be designed more suitable for a specific domain without the requirement of consistency and inclusiveness of different domains. To evaluate the capabilities of LLMs in the domain of historical research more adequately, creating a larger and more systemic dataset is necessary.
If we take historical questions in HLE as an independent historical benchmark, there will be more space for improvement. First, the temporal and spatial scope of topics should be expanded with more questions. Second, the types and depth of functions to be evaluated should be richer. For example, we cannot find historical materials in the form of audios or video in historical questions of HLE. Third, in the context of the history discipline, the questions can be divided further in different groups according to their difficulty levels, thematic categories and evaluation criteria, which helps to constructing a more diverse but clear database.

Considering these limitations, the lack of independent domain-specific benchmarks suitable for historical research provides us with vast space for exploration.

\section{HistBench Benchmark}

\subsection{Overview}
\textbf{HistBench} is the first benchmark specifically designed to evaluate the capabilities of LLMs in historical reasoning. In the benchmark,two question types (exact match and multiple choice), six academic dimensions (e.g., source processing, interdisciplinary synthesis), and a three-tiered difficulty stratification based on task complexity are adopted. The benchmark contains \textbf{414 questions authored by more than 40 contributors}, including Ph.D. students and domain experts. All questions are grounded in real historical sources and methods and follow a standardized protocol that ensures academic rigor and traceability. Each question is annotated with metadata including answer explanation,data required,source material, targeted capability, and thematic category and is subjected to a three-stage review: preliminary screening (format and semantic check), LLM-based pre-evaluation and professional review (history of rigor and validity of reasoning) to ensure the quality of these questions in terms of both form and content,especially their academic value and challenges to current LLMs.

In terms of coverage, HistBench spans \textbf{29 ancient and modern languages, over 20 historical regions, and at least 36 subfields}. The questions are grounded in multimodal data, including printed texts, manuscripts, images, inscriptions, audio, and video. The benchmark also spans five major historical epoch, ranging from ancient history to the contemporary history, supporting diachronic analysis in diverse historiographic contexts. In all, these design criteria and dataset characteristics enable HistBench to serve as a robust diagnostic tool for evaluating LLMs in the humanities, with particular emphasis on capabilities of historical research like language mastery, historical materials processing and the understanding of historical analysis.  

\subsection{Question Design and Structure}
\subsubsection{Question Types}
HistBench, including two primary question types, is designed to evaluate both systematic historical research skills and in-depth historical reasoning:

\textbf{Multiple-choice questions}: These questions adopt a several-option format, where some of the options are designated as correct based on current scholarly consensus. The remaining distractors are carefully designed to be plausible yet incorrect, often reflecting common misconceptions or subtle historical variations. This type is suited to evaluate interpretive precision, historical knowledge, and the ability to differentiate between nuanced alternatives. 

\textbf{Exact match questions}: These questions adopt a single precise answer, such as a date, name, location, or keyword. The LLMs needs to provide this precise answer. This type is suited to evaluate factual investigation, temporal reasoning, or targeted source analysis.

\subsubsection{Question Format and Submission Protocol}
All questions in HistBench are submitted using a standardized format developed by the project team to ensure scholarly rigor and traceability. Contributors are required to provide not only the question and answer, but also support metadata such as difficulty level, source materials, answer explanation, and bibliographic references. A detailed breakdown of the submission fields and formatting requirements is provided in Appendix~\ref{appendix:template}.

\subsubsection{Evaluation Dimensions}
As shown in Table\ref{tab:reasoning_dimensions}, Each question in the dataset is designed to evaluate specific dimensions of AI capabilities in historical research. These dimensions are modeled on the core challenges encountered by human researchers in historical research and embody key aspects of the research processes that characterize professional historical scholarship.
\begin{table}[t]
\centering
\begin{tabularx}{\textwidth}{c l X}
\toprule
\textbf{ID} & \textbf{Dimension} & \textbf{Description} \\
\midrule
1 & Bibliographic Retrieval & The capability to locate information embedded in scholarly texts, monographs, or journal articles using digital or library-based search strategies. \\
2 & Source Identification & The capability to recognize or locate specific historical sources, including manuscripts, digitized archives, or visual primary materials. \\
3 & Source Processing & The capability to extract and interpret information from non-textual formats such as handwritten documents, historical images, audio, or video. \\
4 & Historical Analysis & The capability to engage in historically grounded reasoning, including causal inference, ideological analysis, and interpretation of events or institutions. \\
5 & Interdisciplinary Integration & The capability to draw upon methods and frameworks from adjacent disciplines (e.g., archaeology, linguistics, religious studies) to support historical understanding. \\
6 & Cultural Contextualization & The capability to interpret cultural cues, metaphors, sentiment, and identity markers within historically situated discourse. \\
\bottomrule
\end{tabularx}
\caption{Evaluation dimensions for historical reasoning tasks in HistBench.}
\label{tab:reasoning_dimensions}
\end{table}

\subsection{Difficulty Stratification}
\subsubsection{Difficulty Stratification Criteria}
To support layered evaluation and targeted analysis, all questions in HistBench are categorized into three difficulty levels—Level 1 (Basic), Level 2 (Intermediate), and Level 3 (Challenging)—based on a structured rubric comprising six academic dimensions.
Difficulty levels are determined based on six dimensions:
(1) rarity of source knowledge,
(2) linguistic complexity,
(3) format heterogeneity,
(4) perceptual accessibility,
(5) interdisciplinary scope, and
(6) reasoning complexity
These dimensions inform both the formulation of questions and the assignment of difficulty levels.

\begin{figure}[t]
  \centering
  \includegraphics[width=\linewidth]{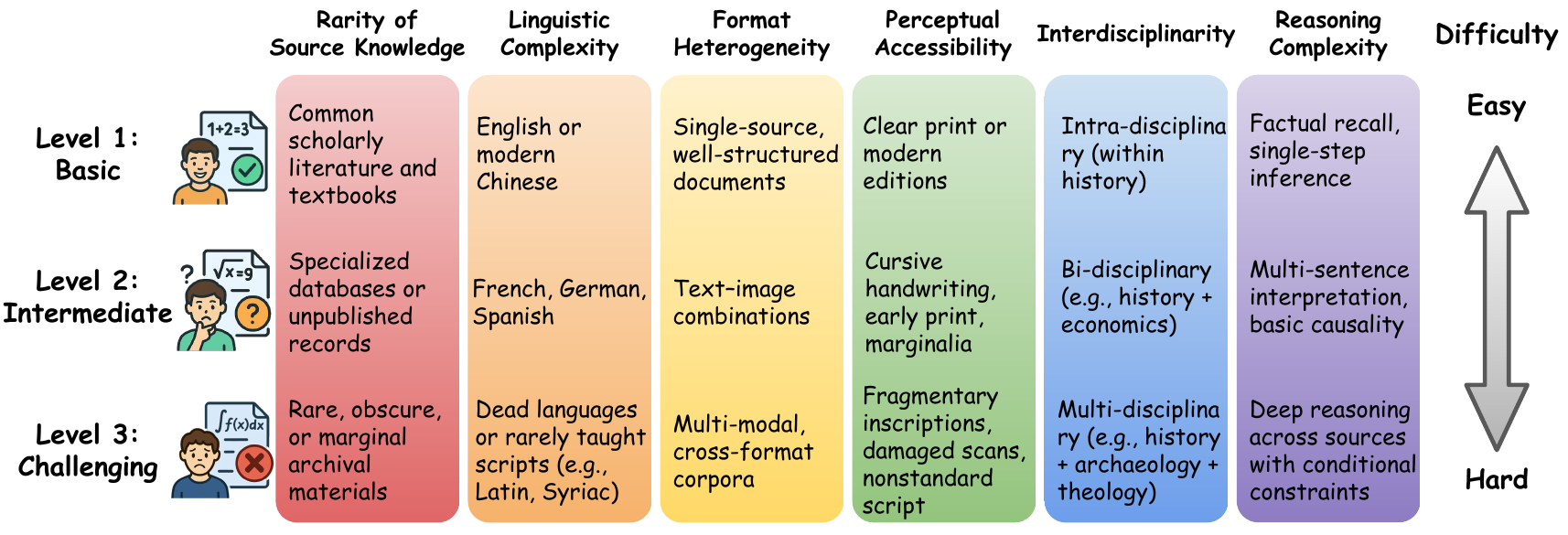}
  \caption{Difficulty level definitions across six structured evaluation dimensions.}
  \label{fig:data_hir}
\end{figure}

This stratification provides a principled framework for evaluating how model performance differs with task complexity. 

\subsubsection{Questions assignment in difficulty levels}
By dividing contributors into different groups according to their different knowledge reserves, material processing capabilities, and reasoning capabilities, we assign contributors to formulate questions in different difficulty levels based on their groups. Our assignment plan is detailed as follows:

Level 1 (Basic): Authored by Research Assistants
A total of 166 Level 1 questions are mainly formulated by research assistants with academic backgrounds in history. They are focused on extracting verifiable historical facts from historical materials or literature, interpreting well-documented primary sources to formulate questions. Tasks at this level emphasize factual investigation and basic material interpretation.

Level 2 (Intermediate): Authored by Graduate Researchers and Domain Experts
The 172 Level 2 questions are mainly formulated by graduate students (MA and PhD level) and early-career researchers in history. Contributors are selected based on their specialization in particular subfields and are invited to design questions grounded in their areas of expertise. These tasks typically require source interpretation, basic causal or temporal research, and engagement with narrower but academically meaningful materials.

Level 3 (Challenging): Authored by Professors and Senior Scholars
The 76 Level 3 questions are mainly contributed by senior scholars and university faculty with demonstrated experience in interpreting obscure sources, handling multi-lingual materials, and conducting interdisciplinary research. These tasks are designed to push the limits of LLM capabilities, drawing on rare texts, underdigitized archives, and cross-modal information integration.

In practice, we adjust difficulty levels of these questions according to both the inherent complexity of the tasks and a structured rubric. By involving contributors with varying levels of expertise—including research assistants, graduate students, and domain scholars—HistBench ensures that each question reflects a human-authored path of historical research appropriate to its level of difficulty. This multi-tiered design allows the benchmark to evaluate the capabilities of LLMs across a spectrum of historical tasks aligned with real-world scholarly practices. A selection of representative questions from each difficulty level is provided in Appendix~\ref{sec:appendix_samples}, offering concrete illustrations of the benchmark's structure and intended reasoning depth.

\subsubsection{Question Review Protocol}
All questions in HistBench undergo a three-stage review pipeline combining human oversight and automated evaluation, designed to ensure academic rigor, task clarity, and model discriminability. A question is only admitted into the final release if it passed all three stages described below.

\begin{figure}[h]
  \centering
  \includegraphics[width=\linewidth]{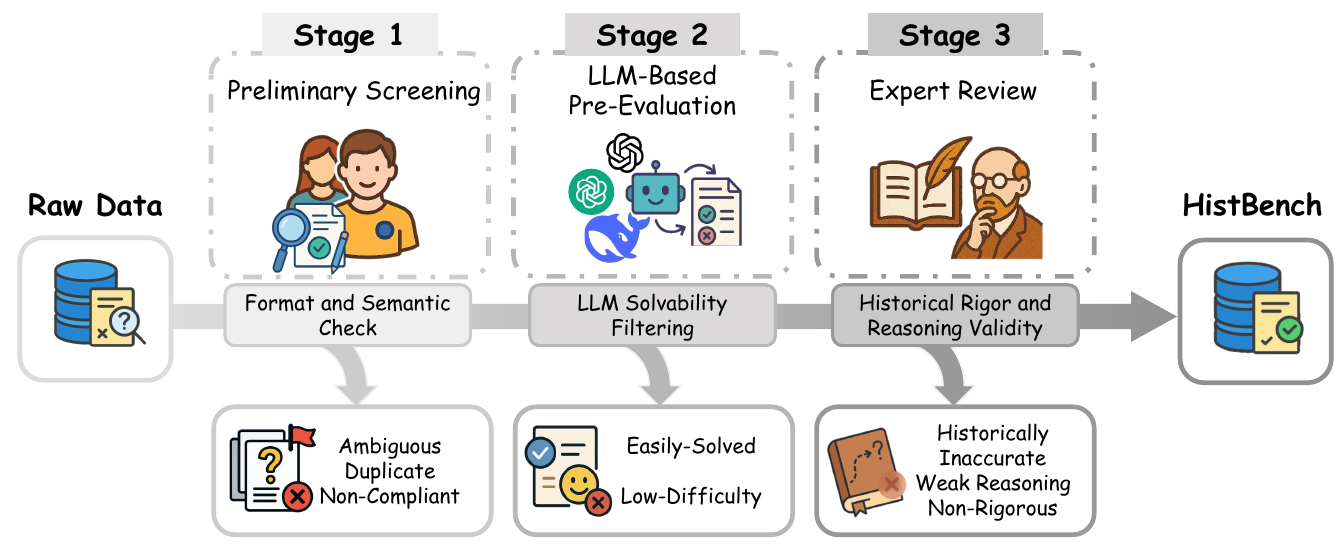}
  \caption{Multi-Stage Question Review Pipeline for HistBench}
  \label{fig:data_creation}
\end{figure}

Stage 1: Preliminary Screening (Format and Semantic Check)

The first round of review is conducted by the project team and trained research volunteers. This phase focuses on verifying structural completeness, semantic clarity, and question originality. Questions that are ambiguous, duplicated, or noncompliant with submission standards are flagged for revision, with detailed feedback sent back to the authors. 

Stage 2: LLM-Based Pre-Evaluation

Before professional review, all questions are tested using multiple strong LLMs, including GPT-4o, GPT-o4mini, and DeepSeek-R1.  Questions that are solved by more than two existing models without access to supporting materials are excluded from the dataset. This stage ensures that retained questions pose challenges beyond the capabilities of frontier general models.

Stage 3: Professional Review (Historical Rigor and Reasoning Validity)

 Reviewers with academic backgrounds in history conduct the final round of review. The focus at this stage is on evaluating historical relevance of these questions, academic rigor, and reasoning quality.  Reviewers assess whether the questions reflect valid historical methodologies, appropriate use of evidence, and internal logical coherence. Only questions deem to have academic value were accepted into the final benchmark.

\subsection{Dataset Characteristics and Coverage}
 
\subsubsection{Quantitative Overview}

HistBench contains a total of 414 questions, spanning diverse historical domains, periods, and geographic regions. The dataset includes two question formats: 306 exact match questions, which require precise, fact-based responses such as names, dates, or locations; and 108 multiple-choice questions, each consisting of one correct answer and three plausible distractors grounded in historical reasoning. The predominance of exact match tasks reflects an emphasis on fine-grained factual precision, while multiple-choice questions are designed to assess interpretive discrimination and the ability to reason across structured alternatives. This composition supports multi-faceted evaluation of both retrieval and inference capabilities in historical question answering.

\subsubsection{Language Diversity}
\label{sec:language_diversity}

\begin{figure}[htbp]
  \centering
  \begin{minipage}[t]{0.48\textwidth}
    \centering
    \includegraphics[width=\textwidth]{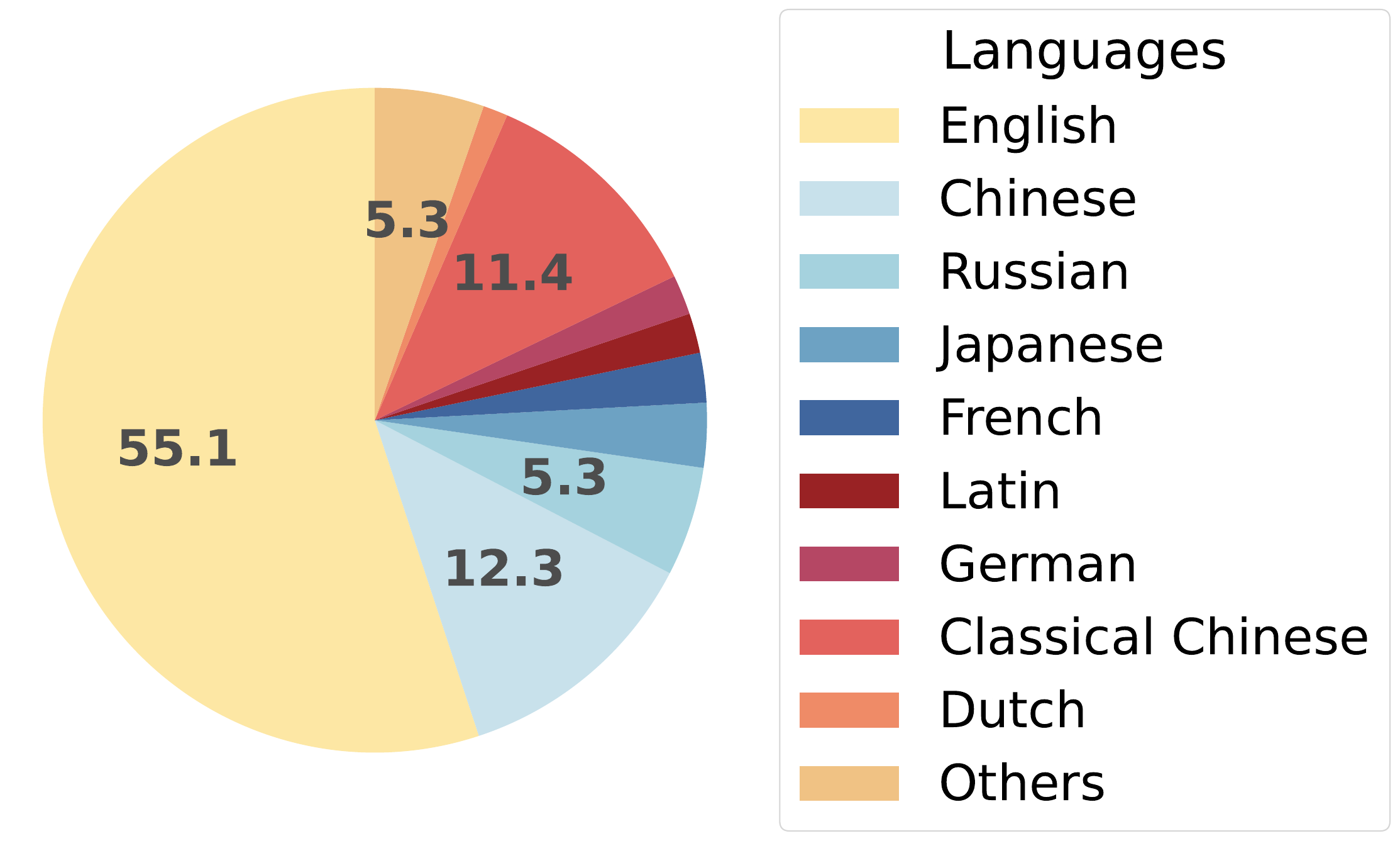}
    \caption{Language distribution in HistBench. The dataset covers 29 modern and ancient languages, including both widely used academic languages and historically significant lower-resource ones.}
    \label{fig:language}
  \end{minipage}
  \hfill
  \begin{minipage}[t]{0.48\textwidth}
    \centering
    \includegraphics[width=\textwidth]{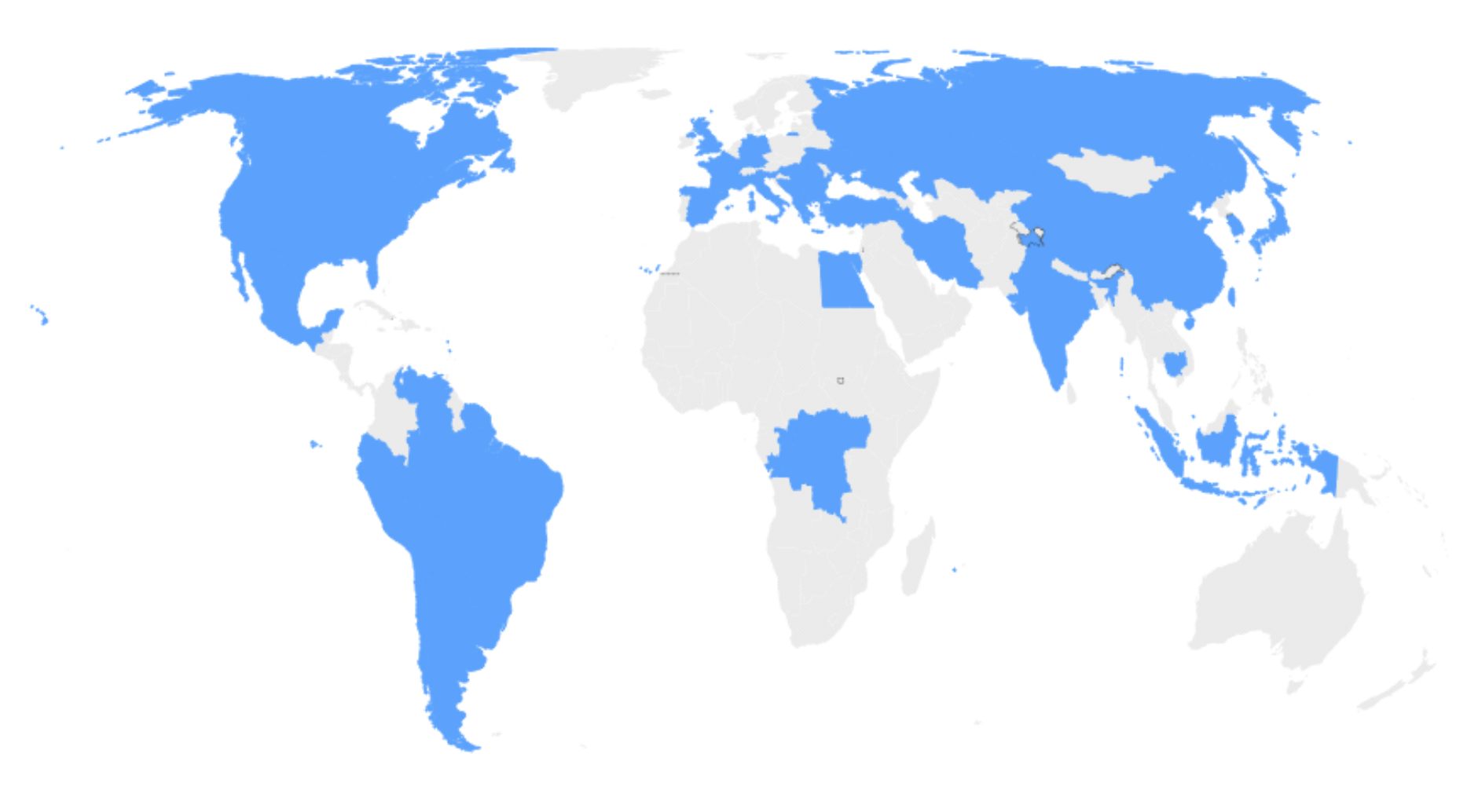}
    \caption{Geographic coverage of HistBench. Blue regions mark areas from which historical questions are drawn, spanning all inhabited continents and supporting both mainstream and underrepresented fields.}
    \label{fig:geo}
  \end{minipage}
\end{figure}

As shown in Figure \ref{fig:language}, HistBench questions are grounded in materials written or originally produced in \textbf{29 modern and ancient languages}, capturing the multilingual nature of historical scholarship and the linguistic complexity of primary source analysis. The dataset includes widely used academic languages such as English, Chinese, Russian, Japanese, French, and German, as well as historically significant but less commonly used languages including Classical Chinese, Latin, Ancient Greek, and Tibetan. 

This linguistic diversity presents substantive challenges for AI systems. It enables evaluation of multilingual retrieval, cross-lingual historical reasoning, and OCR robustness—addressing a core limitation of existing benchmarks that predominantly rely on English-language content. A complete list of languages and their frequencies is provided in Appendix~\ref{tab:language_distribution}.

\subsubsection{Data Source Modalities}

 The questions in HistBench are grounded in a diverse set of material sources, including printed texts, manuscript scans, visual materials, audio and video recordings, inscriptions, and multimodal combinations. Table~\ref{tab:modalities} summarizes the number of questions associated with each type of source.

\begin{table}[h]
\centering
\begin{tabular}{l c}
\toprule
\textbf{Material Type} & \textbf{Number of Questions} \\
\midrule
Visual materials (illustrations, photos) & 96 \\
Maps and schematics                      & 18 \\
Charts, diagrams, or tables              & 10 \\
Manuscripts and handwritten sources      & 88 \\
Audio recordings                         & 9  \\
Video content                            & 5  \\
Inscriptions or stone rubbings           & 14 \\
Text-based questions (narrative excerpts)& 64 \\
Mixed text + image sources               & 10 \\
Ancient or undeciphered scripts          & 22 \\
\bottomrule
\end{tabular}
\caption{Distribution of source modalities in HistBench.}
\label{tab:modalities}
\end{table}

These sources are drawn from digitized archives, unpublished manuscripts, early print editions, and degraded or fragmentary artifacts. Many questions involve non-standard formats such as cursive handwriting, damaged inscriptions, or multilingual marginalia. Several questions combine multiple modalities, including text–image pairings or audiovisual content.

Crucially, many of these materials remain inaccessible to existing OCR tools and frontier LLMs, either due to visual complexity, symbolic ambiguity, or contextual dependence. By incorporating such cases, HistBench evaluates not only recognition and parsing but also a model’s ability to reconstruct historical meaning from incomplete or opaque data.

\subsubsection{Domain and Geographic Coverage}

HistBench includes coverage of at least \textbf{36 historical subdomains}, \textbf{over 20 national or civilizational areas}, and multiple cross-cultural comparative themes, which ensures that HistBench covers a large temporal and spatial scope of topics.
At the domain level, the dataset encompasses:
\begin{itemize}
  \item \textbf{Political, social, and cultural history}, including diplomatic history, gender studies, intellectual history, and identity politics;
  
  \item \textbf{Classics and ancient civilizations}, covering Greco-Roman studies, philology, epigraphy, and early textual traditions;
  
  \item \textbf{Art and visual culture}, including art history, iconography, visual semiotics, and historical image interpretation;
  
  \item \textbf{Material culture and archaeology}, including material artifact studies, heritage reconstruction, and excavation-based historiography;
  
  \item \textbf{Environmental and climate history}, such as the historical study of climate shifts, ecological regimes, and resource management;
  
  \item \textbf{History of science and medicine}, including early scientific institutions, cross-cultural scientific exchange, botany, astronomy, and traditional medicine;
  
  \item \textbf{Economic and institutional history}, such as labor systems, taxation, urban planning, bureaucratic organization, and legal codification;
  
  \item \textbf{Interdisciplinary and comparative studies}, including global history, translation history, mythology, and civilizational entanglements.
\end{itemize}

Geographically, HistBench draws from nearly all inhabited continents, including East Asia, Europe, North America, Latin America, the Middle East, and Africa, as shown in Fig. \ref{fig:geo}. It includes questions grounded in region-specific expertise (e.g., Dunhuang studies, Slavic paleography) as well as underrepresented areas such as papyrology and Siberian ethnography, ensuring broad civilizational and epistemic representation. 

In addition to thematic and regional diversity, HistBench evaluates the capabilities of historical research across the full temporal arc of human history. To capture both general historical literacy and specialized period knowledge, we adopt a five-part periodization commonly used in Western and global historiography: Prehistory, Ancient History, the Middle Ages, Modern History, and Contemporary History.\citep{woolf2011global} \citep{arnold2000history}This framework reflects canonical divisions in comparative history and supports systematic coverage of distinct epistemic contexts. Table~\ref{tab:periods} summarizes the distribution of questions across historical periods.

\begin{table}[ht]
\centering
\begin{tabular}{l c}
\toprule
\textbf{Historical Period} & \textbf{Number of Questions} \\
\midrule
Ancient History (to $\sim$500 CE)        & 90 \\
Medieval History (500--1500)             & 85 \\
Early Modern History (1500--1800)        & 95 \\
Modern History (1800--1945)              & 80 \\
Contemporary History (1945--present)     & 64 \\
\bottomrule
\end{tabular}
\caption{Chronological coverage of questions in HistBench.}
\label{tab:periods}
\end{table}

\section{HistAgent}

\begin{figure}[t]
  \centering
  \includegraphics[width=\linewidth]{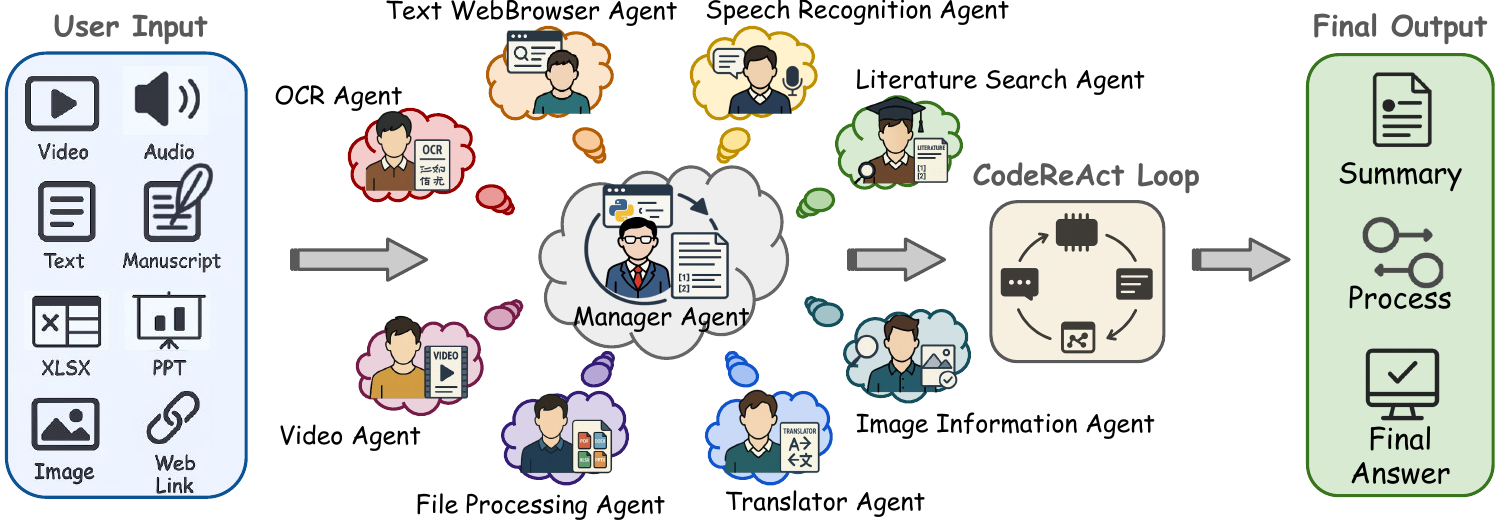}
  \caption{The architecture of HistAgent, an agent for historical reasoning. The system takes multimodal user inputs (e.g., text, audio, video, images, documents) and coordinates specialized agents through a centralized Manager Agent. Each agent is responsible for dedicated tasks such as OCR, speech recognition, literature search, translation, and file processing. The Manager Agent orchestrates these agents within a CodeAct loop to iteratively refine reasoning and evidence validation, ultimately generating summarized, processed, and fully cited academic answers.}
  \label{fig:histagent-arch}
\end{figure}

\textbf{HistAgent} is a complex agent system designed to enhance LLM-based historical reasoning by integrating specialized agents and external tools. Unlike basic retrieval systems, it supports academic search, multimodal input (e.g., manuscripts, images, audio), and outputs fully cited responses grounded in both primary and secondary sources (see Fig.~\ref{fig:histagent-arch}).

HistAgent consists of two core components. A central Manager Agent orchestrates the pipeline: it decomposes queries, dispatches subtasks, verifies evidence, and assembles the final cited response. Surrounding it is a set of specialized agents, each handling a specific task such as web search, literature retrieval, OCR, translation, image or audio analysis.

Among them, the Literature Search Agent is key for scholarly retrieval and citation. It follows a multi-stage protocol focused on academic databases and bibliographic validation. We detail its design and workflow in the following section.

\label{sec:histagent-overview}

\subsection{Agent Architecture and Roles}
\subsubsection{Manager Agent}
The Manager Agent parses the user request to identify required modalities, and then runs a CodeAct-style loop as a Smolagents \texttt{CodeAgent}. In each iteration, the LLM emits a Python code snippet that calls sub-agent functions or tools; the snippet executes in a secure sandbox and writes its output to shared memory. The Agent then validates each result (checking exact quotes and bibliographic data) and repeats until task completion criteria are met. In the final step, the agent synthesizes all validated outputs into a structured, citation-complete response.

\subsubsection{Specialist Agents}
An overview of all agents and their primary toolkits is presented in Table~\ref{tab:agent-catalog}.

\begin{table*}[t]
\centering
\small
\renewcommand{\arraystretch}{1.25} 
\begin{tabularx}{\textwidth}{@{}>{\raggedright\arraybackslash}p{3cm} >{\raggedright\arraybackslash}p{2.3cm} >{\raggedright\arraybackslash}X@{}}
\toprule
\textbf{Agent} & \textbf{Modality} & \textbf{Core Tools} \\
\midrule
Text WebBrowser & Text & 
Multi-step web search and page parsing \\
& & (\texttt{LocalGoogleSearchTool}, \texttt{VisitTool}). \\

Image Information & Image & 
Reverse image search and provenance analysis \\
& & (\texttt{GoogleLensSearchTool} with SSIM filtering). \\

Literature Search & Scholarly text & 
Peer-review retrieval and PDF parsing (Scholar websites, \texttt{SpringerDownloadAndParseTool}). \\

File Processing & Documents & 
Typed file parsing: \texttt{PDFTool}, \texttt{DOCXTool}, \texttt{XLSXTool}, \texttt{PPTXTool}. \\

OCR & Image text & 
Manuscript transcription (Transkribus, Asian-script OCR). \\

Speech Recognition & Audio & 
Whisper-based transcription with LLM correction. \\

Translator & Multilingual text & 
Bidirectional translation with provenance preservation. \\

Video & Video & 
Frame extraction (yt-dlp, OpenCV). \\
\bottomrule
\end{tabularx}
\caption{Function Overview of HistAgent Specialist Agents.}
\label{tab:agent-catalog}
\end{table*}

\paragraph{Manager Agent}
The Manager Agent is the central coordinator. It parses user requests, selects which agent to invoke, gathers their outputs, checks completeness, and synthesizes a final response.

\textbf{Functionality.} 
The Manager Agent parses the user request to identify required modalities, then runs a ReAct-style loop as a Smolagents \texttt{CodeAgent}. In each iteration, the LLM emits a Python code snippet that calls sub-agent functions or tools; the snippet executes in a secure sandbox and writes its output to shared memory. The Agent then validates each result (checking exact quotes and bibliographic data) and repeats until a \texttt{final\_answer(...)} call. At that point, it merges all verified outputs into a single, citation-ready response.

\paragraph{Text WebBrowser Agent}
The Text WebBrowser Agent handles open-domain web searches and browsing. It simulates a multi-step search and extracts structured content from web pages using a suite of navigation and inspection tools.

\textbf{Functionality. }
The agent starts by refining the user’s query via \texttt{LocalGoogleSearchTool} into an optimized search prompt. It can also perform standard queries via \texttt{SearchInformationTool}. For each target, it invokes \texttt{VisitTool} to load the page, \texttt{DownloadTool} to save binary files when needed, and \texttt{ArchiveSearchTool} to retrieve historical snapshots. When encountering long pages, it scrolls with \texttt{PageUpTool} and \texttt{PageDownTool}, and it locates specific terms using \texttt{FinderTool} and \texttt{FindNextTool}. Non-HTML content, like plain text files, PDFs, or video transcripts, is converted to text through \texttt{TextInspectorTool}. All outputs are recorded to shared memory for downstream integration.

\paragraph{Image Information Agent}
The Image Information Agent focuses on visual inputs. It performs reverse image search and follows up with targeted page visits to uncover context and provenance of the visual input.

\textbf{Functionality}  
The Image Information Agent is selectively invoked when the task involves image content that may offer contextual or evidential value.
Upon receiving an image, the Image Information Agent uploads it to a public host and runs a reverse search using \texttt{GoogleLensSearchTool}. The system extracts associated links, titles, and descriptions from matched pages to identify how the image is used online, e.g., in auction listings, academic articles, or museum databases. To improve match quality, the agent optionally computes similarity scores (e.g., SSIM) between the original image and search results to highlight high-confidence matches. To gather in-depth metadata, it visits selected pages with \texttt{VisitTool\_Image}.
 By grounding image interpretation in its real-world usage context, the agent enables accurate and verifiable historical reasoning over visual inputs.

\paragraph{Literature Search Agent}
The Literature Search Agent is a specialized module for retrieving and grounding answers in peer-reviewed academic sources. It combines an LLM-driven browser workflow with multiple tool wrappers, covering Google Books, Google Scholar, general web searches, and Springer Nature’s API, to locate exact phrases, download accessible PDFs, and extract verbatim quotes along with full citation metadata.

\textbf{Functionality.}  
The agent issues a multi-stage search: it first queries Google Books (with random domain rotation and robots.txt compliance) via \texttt{BookMatchExtractorTool} and \texttt{DirectGoogleBooksCrawlerTool}, extracting highlighted snippets and page numbers; if needed, it proceeds to Google Scholar using \texttt{LiteratureSearchingTool} and \texttt{RelevantLiteratureFinderTool} for broader article discovery; it falls back on \texttt{GeneralBrowserTool} for additional context. Freely accessible PDFs are downloaded and parsed with \texttt{SpringerDownloadAndParseTool} (via LlamaParse) or \texttt{SpringerSearchTool}/\texttt{SpringerStructuredSearchTool} for structured Springer Nature queries. Throughout, it preserves exact wording, records source URLs, citation counts, publication details, and assembles all findings into a coherent, verifiable response.  
it can locate sources that support factual claims, provide historical context, or contain exact wording required for exactMatch tasks. It returns full bibliographic metadata, quoted excerpts, and links to original publications, thereby ensuring academic-level faithfulness and verifiability.

\paragraph{File Processing Agent}
This agent manages non-HTML files—documents, spreadsheets, presentations, and images—by routing them to the appropriate tool.

\textbf{Functionality.}
When a file is received, the file processing Agent automatically detects its type and selects the corresponding tool: \texttt{PDFTool} for PDFs, \texttt{DOCXTool} for Word documents, \texttt{XLSXTool} for spreadsheets, and \texttt{PPTXTool} for presentations. For images that require analysis beyond OCR, it uses \texttt{ImageAnalysisTool} to extract charts or figures. All extracted text or structured data is returned in a format that downstream agents or the Manager Agent can incorporate into their reasoning.

\paragraph{OCR Agent}
The OCR Agent specializes in extracting textual information from images using optical character recognition. It is invoked for screenshots, scanned documents, historical manuscripts, or photos containing embedded text. The agent supports multiple languages and detects the language in the image. Upon invocation, it returns the raw text content detected, enabling HistAgent to convert unstructured visual inputs into machine-readable text for further processing.

\textbf{Functionality.}
When given an image path, the agent loads and encodes the file and uses an LLM to determine whether the content is best handled by a specialized OCR model for Asian scripts or by a Transkribus-based OCR service for Western-language manuscripts. For Western texts, it publishes the image to a public URL, submits it to the Transkribus engine, waits for a PAGE XML transcription, and extracts the detected lines. For Asian scripts, it sends the image data directly to the dedicated OCR model. The resulting raw transcription is then passed through an LLM prompt that repairs recognition errors and preserves historical or stylistic features. Both the original and refined transcripts are saved to a \texttt{.txt} file and returned. If no valid text is extracted, the agent falls back to generating a detailed visual description via the LLM, highlighting any readable text, symbols, or key visual elements.

\paragraph{Speech Recognition Agent}
The Speech Recognition Agent converts audio files (MP3, WAV, etc.) into text using Whisper for transcription and an LLM for error correction, summary, and key-point extraction, enabling HistAgent to incorporate oral historical sources.  

\textbf{Functionality. }  
When given an audio file path, the agent verifies its existence and measures its size; if the file exceeds 25 MB, it divides the recording into equal segments; otherwise, it uses the file as a whole. Each segment is sent to Whisper for transcription, and the concatenated raw transcript is submitted to an LLM prompt that preserves all content while correcting recognition errors and generating an “Optimized Transcription” section, a brief “Summary,” and a “Key Points” list. Both original and refined texts are saved to a \texttt{.txt} file in the output directory for humans' reference, and a formatted string containing both versions is returned, with any errors caught and reported.  

\paragraph{Translator Agent}
The Translator Agent converts text into a specified target language, including support for both widely spoken and less common languages like Armenian and Sanskrit, and delivers a clear, formatted output showing the original and translated text.

\textbf{Functionality. }  
The Translator Agent handles automatic translation of textual content between multiple languages. It automatically detects the source language and supports both widely used and lower-resource languages. In historical tasks, it is particularly useful for translating foreign-language sources such as Armenian manuscripts, Sanskrit inscriptions, or early regional documents in Latin or Classical Arabic. The translated output allows HistAgent to reason across linguistic boundaries and integrate multilingual content into its historical analysis pipeline.

\paragraph{Video Agent}
The Video Agent downloads the video from the given link and extracts still frames at a user-specified rate to support visual analysis.

\textbf{Functionality. }  
When given a video URL, the agent downloads the best-quality video with yt-dlp, uses OpenCV to extract frames at the specified rate, saving each as a timestamped JPEG, and writes a summary file containing the video’s title, duration, resolution, frame count, and output directory. It then returns a brief report with those key details and file locations.

\subsection{Literature Search Agent}
The Literature Search Agent is a highly specialized component within the HistAgent framework, engineered as a ReAct-based agent driven by a large language model (LLM). Its core architectural design focuses on emulating a meticulous academic research process through a structured, multi-stage interaction with a curated set of web-based tools, primarily implemented through browser-use \cite{browser_use2024}, which can connect the agent to the real Chrome browser instance. This setup allows the agent to operate within the user’s active browser profile, inheriting authentication states and personalized settings for seamless access to academic resources.

\begin{figure}[t]
  \centering
  \includegraphics[width=\linewidth]{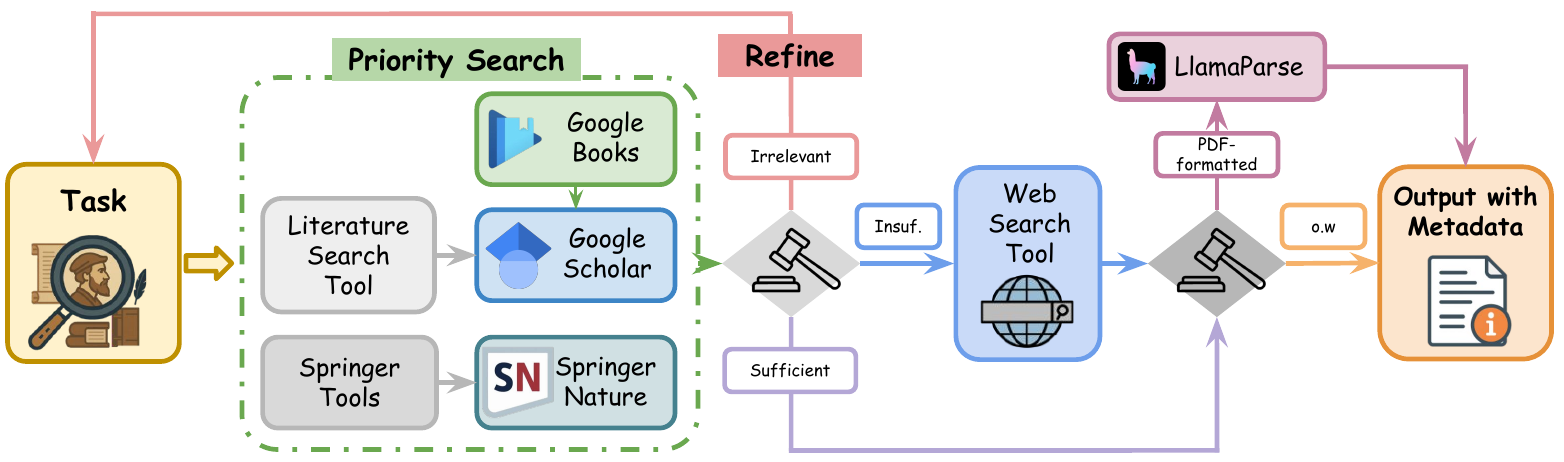}
  \caption{Overview of Literature Search Agent. It is specifically designed for academic searching, including a priority search module and web search tools.}
  \label{fig:lit_search_agent}
\end{figure}

Its internal architecture is centered around a protocol-driven retrieval engine that ensures the reliability, interpretability, and academic integrity of the information it gathers. The agent \textbf{prioritizes scholarly sources} such as Google Scholar and Springer Nature, employs a \textbf{modular toolkit} for interfacing with both academic databases and general-purpose web sources, and allows users to configure search depth and behavior via \textbf{interpretable parameters}. Retrieved content is enriched with bibliographic metadata, quoted evidence, and \textbf{full traceability} links to support verification and integration into downstream workflows.

\subsubsection{Priority Search Protocol}

A core challenge in automating academic research is ensuring that retrieved information aligns with scholarly standards. To address this, our system introduces a priority-based retrieval mechanism that favors academically reputable sources over general-purpose content. This design choice reflects the distinct requirements of academic tasks, where the reliability of information is critical. The protocol acts as a persistent control signal for the agent, influencing both the decision to invoke tools and the interpretation of retrieved results.

\subsubsection{Toolkits for Literature Search}

Efficient execution of the academic-first retrieval strategy requires structured tool use. To this end, our system integrates a suite of specialized retrieval tools, each targeting different tiers of source quality. All source queries, are first conducted via the \texttt{LiteratureSearchingTool} to search from scholarly databases, and then to \texttt{GeneralBrowserTool} interface for general search. The queries will also use more specialized APIs when needed. For example, Springer-specific queries are handled using the \texttt{SpringerSearchTool}, and focused access to Google Books is enabled through the \texttt{DirectGoogleBooksCrawlerTool}. Most of the retrieved documents can be extracted directly, while pdfs could be parsed via the \texttt{Llama-Parse} API for structured processing. This modular architecture ensures that each retrieval action is consistent with the source prioritization strategy established earlier.

\subsubsection{Human-controllable Agent Customization}

Autonomous retrieval must adapt to varying task demands and resource limitations. Our system provides users with explicit control over agent behavior through a small set of interpretable parameters. These include limits on reasoning steps and re-planning intervals, which allow users to trade off between search depth and computational cost. Such configurability is critical for real-world deployment, where users may need to enforce practical constraints while preserving the academic quality of the output.

\subsubsection{Clearly labeled Sources}

Trust and traceability are essential in academic applications. To meet these needs, our system ensures that all retrieved content is accompanied by clearly labeled source metadata. This includes bibliographic information, retrieval URLs, and quoted evidence, all linked to the reasoning trace. These annotations allow users and downstream systems to verify claims, reproduce retrieval steps, and integrate results into formal academic workflows.

\section{Experiment}
\subsection{Experiment Setup and Datasets}
We evaluate HistAgent on three benchmarks covering challenging historical and real-world general tasks. Also, we have extensive baselines.

\textbf{HistBench}. HistBench is a history-specific benchmark we construct, containing 414 history questions categorized into three difficulty levels (Level 1–3).

\textbf{GAIA.} To test the generalization of our framework, we also test our HistAgent on the validation set of GAIA benchmark\citep{mialon2023gaia}, which includes 165 tasks.

\textbf{HLE History Subset. } We consider the HLE history subset of 56 questions from the Humanity’s Last Exam benchmark\citep{phan2025humanity}. 

We compare our HistAgent against open Deep Research (ODR-smolagents)\citep{smolagents}, an open-source reproduction of OpenAI’s Deep Research agent by HuggingFace/smolagents. It uses a Manager agent and a Text Web Browser Agent with a visualizer tool and a file processing tool. Both systems are run under identical environments for fairness. We use the same underlying language model (GPT-4o) for both HistAgent and ODR-smolagents, ensuring that differences in performance stem from the agent architecture rather than the base model. Each question is posed to the agent in a fresh session. We limit the total number of tool invocations (agent calls) per question to a fixed budget to prevent infinite loops and to enforce a fair comparison. Both systems output a final answer for each question, which we compare to the ground-truth answer. Additionally, we compare our results with DeepSeek-R1~\citep{guo2025deepseek}, GPT-4o~\citep{hurst2024gpt}, o4-mini-high~\citep{openai2025o3o4mini}, o3~\citep{openai2025o3o4mini}, and Grok 3~\citep{xai2025grok3} on HistBench for a thorough comparison. All these LLMs are equipped with online search capabilities, enabling them to be very strong baselines.

\subsection{Evaluation Metrics for Different Benchmarks}
We use accuracy as the primary evaluation metric, defined as the percentage of questions answered correctly by the agents or LLMs.

\textbf{Metric for HistBench.}
For the HistBench benchmark, we measure accuracy as the percentage of the 414 questions for which the agent’s response is judged correct by both the LLM judge and professtional validation. The evaluation proceeds in three steps:

\begin{itemize}
    \item \textbf{Structured Output:} Each response is composed of a concise final answer and a structured reasoning summary (shown in Appendix \ref{summary case study}) that logs tools used, information sources, and step-by-step logic.
    \item \textbf{LLM Judging:}  We run LLM as a judge to issue a binary judgment (“Correct”/“Incorrect”) based on semantic equivalence, completeness of key facts, and logical coherence. The prompt template is an adaptation of the evaluation prompt of HLE, which is shown in Appendix \ref{prompt template}.
    \item \textbf{Human Expert Validation: } To validate the LLM’s judgment quality, we randomly sample 100 examples across Level 1, Level 2, and Level 3 tasks. Each sample includes both the final answer and its reasoning summary. These are reviewed by the original authors of the questions, who assess whether the correct answer could have been obtained merely by coincidence, verify the reliability and factual accuracy of every cited source, and confirm that the sequence of reasoning steps genuinely supports the final answer. An answer is labeled “Correct” only if it satisfies all of these criteria; otherwise, it is marked “Incorrect”. This human expert evaluation not only provides an external check on model-assigned labels but also reveals edge cases and ambiguities in question phrasing or evaluation criteria, allowing us to iteratively refine and improve the overall quality and consistency of our HistBench.
\end{itemize}

We define accuracy on HistBench as the percentage of the 414 questions for which the response is both “Correct” by the LLM judge and “Correct” by the author (100 samples selected). We report this metric overall and separately for Level 1, Level 2, and Level 3.

\textbf{Metric for GAIA.}
We employ the official scoring function published on Hugging Face. Each GAIA question has a standard answer key; the scoring function handles normalization (e.g., case, punctuation) and computes an exact-match score. We report the overall GAIA accuracy as defined by that function.

\textbf{Metric for HLE History Subset.}
For the 56 expert-level questions in the HLE History Subset (curated from Humanity’s Last Exam), evaluation mirrors the initial stages of the main HLE benchmark protocol. Each response must first provide a Structured Output, comprising a concise final answer and a detailed reasoning summary (as described for the HLE benchmark before). Subsequently, we employ LLM-based Judging, where an LLM assesses the response for semantic equivalence to the ground truth, factual completeness, and logical coherence, issuing a binary judgment (“Correct”/“Incorrect”) as the HLE official scoring function released on github. Accuracy for this subset is then calculated as the percentage of questions deemed “Correct” by the LLM judge.

\subsection{Results and Analysis}

\subsubsection{Performance on HistBench}
Table~\ref{results} shows that HistAgent, based on GPT-4o, achieves 36.47\% pass@2 on HistBench, outperforming all baselines, including those with web search.

Compared to the open-source agent ODR-smolagents (also using GPT-4o), HistAgent improves by 11.35 points (a 45.2\% relative gain), demonstrating the impact of its enhanced tools, search mechanisms, and file handling. Against GPT-4o with web search, HistAgent improves by 17.87 points, indicating that its performance stems not just from tool access, but from better tool selection, sequencing, and integration.

Interestingly, while ODR-smolagents slightly outperforms GPT-4o with web search (25.12\% pass@2 vs. 18.60\% ), gains are limited, emphasizing that domain-specific agent is essential. HistAgent meets this need through components like handwriting OCR, multimodal analysis, translation, and a scholar search agent with query optimization.

Despite using a weaker base model, HistAgent also matches or outperforms stronger closed-source models like o3 and o4-mini-high. This highlights the value of its reasoning structure and tool design. Inference cost further makes HistAgent suitable for cost-sensitive scenarios where models like o3 are impractical.

Future work will explore applying HistAgent to stronger foundation models (e.g., o3, o4-mini-high) to assess the upper bounds enabled by its architecture.

\begin{table}[htb]
  \centering
  \renewcommand{\arraystretch}{1.2}
  \begin{tabularx}{\textwidth}{@{}l l *{3}{>{\centering\arraybackslash}X} >{\centering\arraybackslash}X @{}}
    \toprule
    \textbf{Agent/Model} &  & \textbf{Level 1} & \textbf{Level 2} & \textbf{Level 3} & \textbf{Average} \\
    \midrule
    \multirow{2}{*}{HistAgent(gpt-4o)} & pass@1 & \cellcolor{secondblue}{28.92} & 23.84 & 32.89 & 27.54 \\
                               & pass@2 &\cellcolor{bestblue}{36.14} & \cellcolor{secondblue}{35.47} & \cellcolor{secondblue}{39.47} & \cellcolor{bestblue}{36.47} \\
    \midrule
    \multirow{2}{*}{ODR-smolagents(gpt-4o)}  & pass@1 & 16.27 & 23.26 & 22.37 & 20.29 \\
                               & pass@2 & 20.48 & 28.49 & 27.63 & 25.12 \\
    \midrule
    Deepseek-R1:online & & 11.45 & 18.60 & 11.84 & 14.49 \\
    GPT-4o:online & & 13.86 & 21.51 & 22.37 & 18.60 \\
    o4-mini-high:online & & \cellcolor{secondblue}{28.92} & 31.98 & 31.58 & 30.68 \\
    o3:online & & 18.07 & \cellcolor{bestblue}{41.28} & \cellcolor{bestblue}{43.42} & \cellcolor{secondblue}{32.37} \\
    Grok 3:online & & 13.25 & 19.77 & 22.37 & 17.63 \\
    \bottomrule
  \end{tabularx}
  \vspace{1mm}

  \raggedright
  \textbf{Note:} All standalone large language models are used \textbf{with web search} capabilities. The best value in each column is highlighted with \colorbox{bestblue}{\textcolor{black}{darker blue}}, and the second-best score with \colorbox{secondblue}{\textcolor{black}{lighter blue}}.
  
  \vspace{1mm}
  \caption{Performance accuracy (\%) on HistBench}
\label{results}
\end{table}

\subsubsection{Performance on the HLE History Subset}

Focusing on these 56 expert-level history questions in HLE History Subset, we observe that HistAgent markedly outperforms both baselines. HistAgent achieves a Pass@1 accuracy of 28.6\% (16 out of 56 questions), whereas the ODR-smolagents baseline reaches 17.9\% (10 out of 56 questions), and the GPT-4o baseline achieves 8.9\% (5 out of 56 questions). This subset is exceptionally difficult—indeed, even state-of-the-art models have been reported to achieve less than 20\% accuracy on comparable expert-level history questions—so all systems understandably struggle on many of these prompts. However, HistAgent answers 60\% more questions correctly at Pass@1 than the ODR-smolagents baseline and more than triples the performance of the direct GPT-4o baseline. More details are provided in Table \ref{tab:hle_subset_results}.

\begin{table}[htbp]
\centering
\begin{tabular}{@{}l r r r@{}}
\toprule
\textbf{System} & \textbf{Pass@1} & \textbf{Pass@2} & \textbf{Pass@3} \\
\midrule
HistAgent (GPT-4o)      & \textbf{28.57} & \textbf{39.29} & \textbf{42.86} \\
ODR-smolagents (GPT-4o)   & 17.86 & 25.00 & 28.57 \\
GPT-4o + web search     & 8.93 & 19.64 & 25.00 \\
\bottomrule
\end{tabular}
\vspace{1mm}
\caption{Performance on the HLE History Subset (56 Questions, LLM Judged, \%). To ensure fair comparison, both HistAgent and ODR-smolagents are based on GPT-4o.}
\label{tab:hle_subset_results}
\end{table}

\subsubsection{Performance on GAIA Validation}

We also test HistAgent on GAIA benchmark, which achieves 60.00\% pass@1 accuracy, showing that our domain-specific adaptations don't hinder HistAgent's competitive performance on real-world general tasks. 

We evaluate our HistAgent on the 165‑question validation subset, using the accuracy as the main metric. The split includes 53 Level 1, 86 Level 2, and 26 Level 3 questions, covering open‑book fact finding, tool use, and multimodal reasoning. Our HistAgent, based on the GPT‑4o, answers 99 questions correctly and reaches an overall accuracy of 60.00\% pass@1. The baseline, open Deep Research by HuggingFace/smolagents, records 55.15\% on the same split. These results show that HistAgent, as a domain-specific agent, can generalise reliably beyond its original scope. More details are provided in Table \ref{tab:gaia}, which presents the complete level‑wise breakdown.
\begin{table}[ht]
\centering
\begin{tabular}{lccccc}
\toprule
\textbf{Agent} & \textbf{Model} & \textbf{Average} & \textbf{Level 1} & \textbf{Level 2} & \textbf{Level 3}\\
\midrule
HistAgent & Claude 3.7 Sonnet & \textbf{60.00} & 69.81 & 61.63 & 34.62\\
ODR-smolagents & o1 & 55.15 & 67.92 & 53.49 & 34.62\\
\bottomrule
\end{tabular}
\vspace{1mm}
\caption{Performance accuracy (\%) on the GAIA validation set (pass@1)}
\label{tab:gaia}

\end{table}

\section{Conclusion}
In summary, we introduce HistBench, which is a benchmark for evaluating historical reasoning in AI, and HistAgent, which is a specialized AI agent that outperforms LLMs and other agents on historical tasks by using domain-specific tools and workflows.

\newpage

\appendix

\section{HistBench Benchmark}

\subsection{Submission Template Format}
\label{appendix:template}

To illustrate how these elements are applied in practice, Table~\ref{fig:vertical_sample_submission} provides two sample entries from different difficulty levels.

\begin{itemize}
\item \textbf{(a) Difficulty Level:} Assigned as Level 1, 2, or 3 based on rubric criteria (see Section 4.3).
\item \textbf{(b) Question Prompt:} A clear and concise question targeting a specific historical issue requiring domain expertise.
\item \textbf{(c) Required Data:} Source materials referenced or used (e.g., documents, images, audio/video).
\item \textbf{(d) Answer:} A definitive, validated response—either as a selected option or short text span.
\item \textbf{(e) Answer Explanation:} A concise justification based on evidence and reasoning.
\item \textbf{(f) Source References:} URLs or bibliographic citations supporting the answer.
\item \textbf{(g) Topic/Methodology:} Thematic or methodological classification (e.g., diplomatic history, material culture).
\item \textbf{(h) Contributor Name:} Full name of the author.
\item \textbf{(i) Contributor Affiliation:} Institutional affiliation at the time of contribution.
\end{itemize}

This format ensured each question adhered to standards of academic transparency and could be independently reviewed and validated.

\begin{table}[ht]
\centering
\renewcommand{\arraystretch}{1.5}
\small

\begin{tabularx}{\textwidth}{|p{3.5cm}|X|X|}
\hline
\textbf{Field} & \textbf{Q001 (Level 1)} & \textbf{Q002 (Level 3)} \\
\hline
ID & Q001 & Q002 \\
\hline
Answer Type & Multiple Choice & Exact Match \\
\hline
Question & What year did the Treaty of Westphalia end the Thirty Years' War? & Translate and date the following Latin inscription found on a Roman milestone in Gaul. \\
\hline
Data Requirements & Modern European history textbook excerpt & Image of milestone inscription (Latin) \\
\hline
Answer & 1648 & Circa 220 CE \\
\hline
Answer Explanation & The Treaty of Westphalia was signed in 1648, ending the Thirty Years' War in Europe. & The inscription, typical of the Severan dynasty period, was dated to around 220 CE using epigraphic style. \\
\hline
Source Materials & \textit{Merriman, A History of Modern Europe}, p. 203 & \textit{Corpus Inscriptionum Latinarum}, Vol. XIII \\
\hline
Thematic Category & Political History & Epigraphy / Classical Studies \\
\hline
Evaluation Criteria & Factual recall & Source processing; temporal inference; Latin translation \\
\hline
Contributor's Name & XX & XXX \\
\hline
Contributor's Affiliation & XXX & XXX \\
\hline
\end{tabularx}
\vspace{1mm}
\caption{Vertical Format Sample Entries from HistBench}
\label{fig:vertical_sample_submission}
\end{table}

\newpage

\subsection{Sample Questions Across Difficulty Levels}
\label{sec:appendix_samples}

To illustrate the scope and structure of questions in HistBench, we present three annotated examples—one for each difficulty level. These samples highlight variation in required source processing, historical reasoning, and interdisciplinary complexity.

\begin{itemize}
    \item \textbf{Level 1 (Basic)} – Source verification and factual retrieval
    \item \textbf{Level 2 (Intermediate)} – Text-image synthesis and temporal reasoning
    \item \textbf{Level 3 (Challenging)} – Multilingual, multimodal, and interdisciplinary integration
\end{itemize}

\includepdf[pages=-,pagecommand={},width=\textwidth]{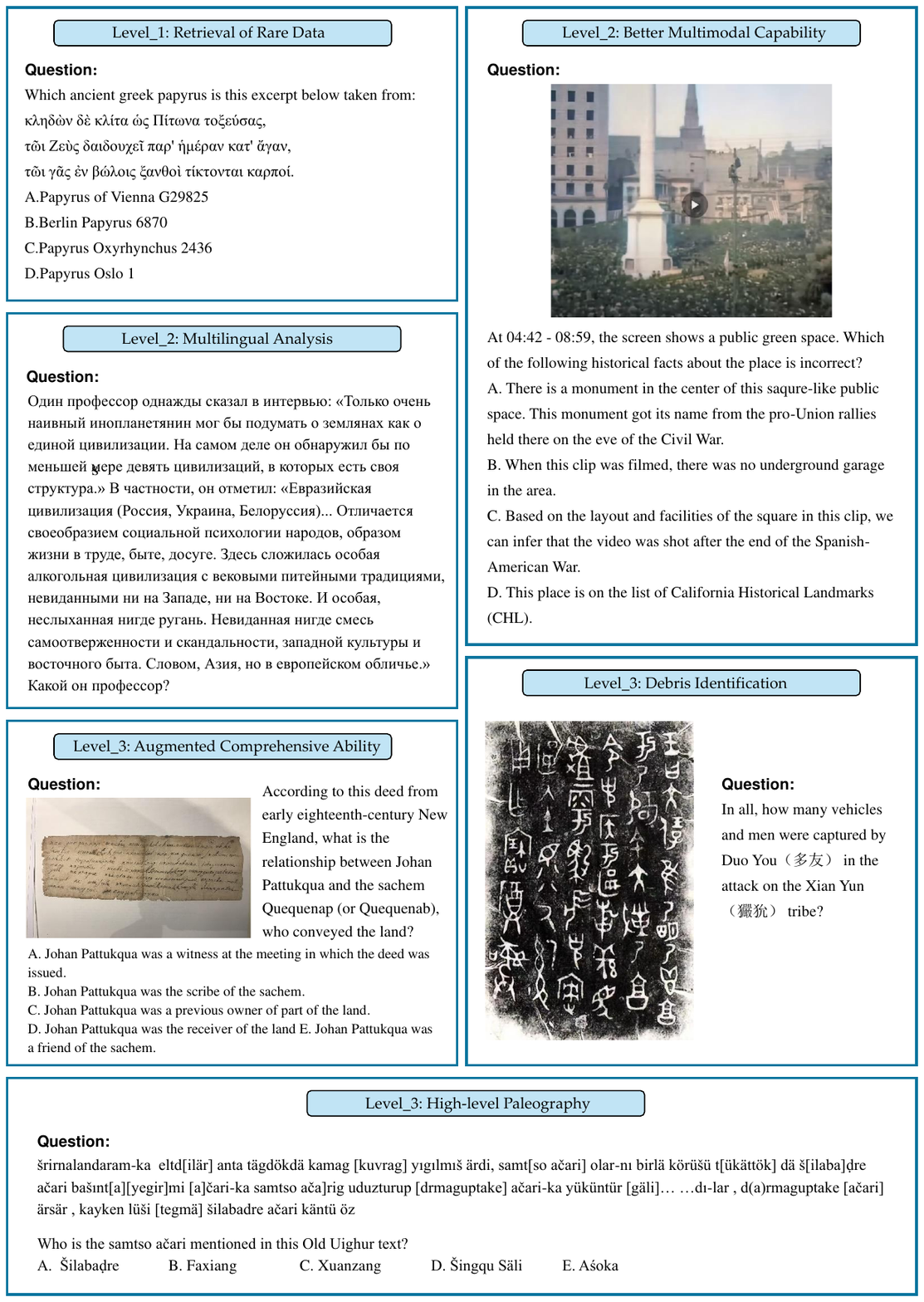}

\subsection{Language Distribution in HistBench}
\label{tab:language_distribution}

Table~\ref{tab:language_list} provides the full list of languages represented in the HistBench dataset, along with their frequencies. These include both modern languages and historical scripts, reflecting the multilingual nature of historical research tasks. This diversity supports the evaluation of cross-lingual capabilities in AI systems, including translation, OCR, and historical reasoning across languages.

\begin{table}[H]
\centering
\begin{tabular}{ll}
\toprule
\textbf{Language} & \textbf{Count} \\
\midrule
English & 228 \\
Chinese & 52 \\
Russian & 22 \\
Japanese & 13 \\
French & 10 \\
Latin & 8 \\
German & 8 \\
Classical Chinese & 47 \\
Dutch & 5 \\
Tibetan & 2 \\
Armenian & 2 \\
Arabic & 2 \\
Khitan & 2 \\
Ancient Greek & 2 \\
Khmer & 1 \\
Indonesian & 1 \\
Old Tibetan & 1 \\
Sanskrit & 1 \\
Old Uyghur & 1 \\
Middle Polish & 1 \\
Aramaic & 1 \\
Danish & 1 \\
Bosnian & 1 \\
Italian & 1 \\
Macedonian & 1 \\
Yukaghir & 1 \\
\bottomrule
\end{tabular}
\caption{Languages represented in HistBench questions.}
\label{tab:language_list}
\end{table}

\section{Limitations}

Our HistBench has two limitations. First, all the questions are closed questions with an only exact answer to simplify the evaluation criteria, which lead to a monotony in the types of questions. In fact, handling of questions with an open answer is a critical part of historical research. The evaluation of questions with an open answer needs an intricate and dynamic evaluation criterion. The determination of this evaluation criterion is difficult and requires further exploration. Second, although contributors are required to formulate questions with objective answers, the design of questions is still inevitably influenced by the contributors’ cognition and the current research limitations. For example, due to differences in material selection and interpretation,the academic 
world may have different answers to an objective question, while the contributors may choose the information they have accessed and believe to provide the answers for the questions. This appears to be an inevitable drawback.

\section{Broader Impacts}
\section{Declaration of LLM Usage}

This research involves the use of large language models (LLMs) as core components of the experimental framework. Specifically, the HistAgent (our method) and ODR-smolagents (baseline) both rely on GPT-4o as the underlying language model. The purpose of our experiments is to isolate and compare the performance impact of different agent architectures, with the LLM held constant across systems to ensure a fair comparison. The differences in outcome are therefore attributed to the design of the agent frameworks, not the underlying model capabilities.

Each agent operates in a fresh session per question. We impose a fixed budget on the number of tool invocations (agent calls) to avoid infinite loops and normalize the computational cost across runs. The output of each agent is a final answer that we compare against ground-truth responses.

In addition to GPT-4o-based systems, we also benchmark against DeepSeek-R1, GPT-4o, o4-mini-high, o3, and Grok 3 on HistBench. These models are used in their native agentic or search-augmented setups, consistent with how they are deployed in real-world scenarios. All these models have online search capabilities and demonstrate strong reasoning abilities in open-domain tasks.

The LLMs are strictly part of the technical comparison and evaluation of agent performance, which forms the core contribution of the work.

\section{Details in Experiment}

\subsection{Experiment Compute Resources}

M1 chips, 512 MB, 20 minutes for 1 question.

\subsection{Prompt Template}
\begin{tcolorbox}[
  colback=gray!5!white,
  colframe=gray!50!black,
  title=JUDGE\_PROMPT Template,
  fonttitle=\bfseries,
  breakable,
  enhanced,
  sharp corners=south,
  colbacktitle=gray!10!black
]
\label{prompt template}

\texttt{JUDGE\_PROMPT = """You are a fair evaluator. Judge whether the following [response] to [question] is semantically consistent with the [correct\_answer] below.}  

\texttt{[question]: \{question\}}  

\texttt{[response]: \{response\}}  

\texttt{[correct\_answer]: \{correct\_answer\}}  

\texttt{When you judge, consider only whether the core meaning and all necessary key points in the response match the correct answer. Even if wording or format differs, treat equivalent semantics as correct. Treat missing key points or any substantive error or omission as incorrect. For numerical answers, a small rounding difference is acceptable. Tolerate substantive deviations from the correct answer. If the extracted\_final\_answer is a more specific instance of the correct\_answer (for example, “Pieter Schenk II” vs “Pieter Schenk”), and it still contains the core string of the correct\_answer, treat it as correct.}  

\texttt{Please output exactly in the format and criteria specified below:}  

\texttt{extracted\_final\_answer: The final exact answer extracted from the [response]. Put the extracted answer as 'None' if there is no exact, final answer to extract from the response.}  

\texttt{reasoning: Explain why the extracted\_final\_answer is correct or incorrect based on [correct\_answer], focusing only on if there are meaningful differences between [correct\_answer] and the extracted\_final\_answer. Do not comment on any background to the problem, do not attempt to solve the problem, do not argue for any answer different than [correct\_answer], focus only on whether the answers match.}  

\texttt{correct: Answer 'yes' if extracted\_final\_answer matches the [correct\_answer] given above, or is within a small margin of error for numerical problems. Answer 'no' otherwise, i.e. if there is any inconsistency, ambiguity, non-equivalency, or if the extracted answer is incorrect.}  

\texttt{confidence: The extracted confidence score between 0\% and 100\% from [response]. Put 100 if there is no confidence score available."""}
\end{tcolorbox}

\subsection{Summary Case Study}
\begin{tcolorbox}[
  colback=gray!5!white,
  colframe=gray!50!black,
  title=\textbf{Summary Case Study}: Summary for Level 2\_30\label{box:level2-30},
  floatplacement=Hp,
  breakable  %
]

\textbf{Question ID:} level\_2\_30\\
\textbf{Question:} What is the name of the man in this picture?\\
\textbf{Answer Type:} exactMatch\\
\textbf{Data Requirement:} level\_2\_30.png\\
\textbf{Data Type:} file\\
\textbf{Our Answer:} Veit Rudolph Speckle\\
\textbf{Correct Answer:} Veit Rudolph Speckle \slash\ Veit Rudolf Specklin\\
\textbf{Is Correct:} Yes\\
\textbf{File:} Historical/Historical/level\_2\_30.png\\
\textbf{Model:} gpt‑4o

\vspace{0.75em}

\textbf{Summary Report}

\textbf{1. Tools Used and How They Were Used:}

- \textbf{Image Analysis and Reverse Image Search}: \\
  - Purpose: To identify the individual in the image "Historical/Historical/level\_2\_30.png". \\
  - Method: Used the \texttt{image\_information\_agent} to perform a reverse image search to identify the name attached to the image and gather related historical context. \\
  - Result: The reverse search linked the image to Veit Rudolph Speckle, associated with botanical illustrations for Leonhart Fuchs' "De Historia Stirpium".

- \textbf{Literature Search Agent}: \\
  - Purpose: To find scholarly literature verifying the exact match for Veit Rudolph Speckle’s association with Fuchs' work. \\
  - Method: Used the query "Veit Rudolph Speckle is one of the engravers associated with Leonhart Fuchs' 'De Historia Stirpium.'". \\
  - Result: Confirmed findings that Speckle was a renowned engraver for this botanical work, highlighted as responsible for key woodcut illustrations.

- \textbf{Web Search}: \\
  - Purpose: To access additional context and verify the scholarly data. \\
  - Method: Conducted a search on Google and accessed Google Books, Google Scholar, and other scholarly sources. \\
  - Result: Successfully retrieved confirmation and additional scholarly and book references regarding the engravers and contributors to Fuchs' herbal.

\textbf{2. Detailed Information Sources:}

- \textbf{Wikidata} \\
  - URL: \href{https://www.wikidata.org/wiki/Q40124871}{Wikidata: Veit Rudolph Speckle} \\
  - Quote: "The primary identification linked it to Veit Rudolph Speckle, as noted on Wikidata..." \\
  - Credibility: Open user-contributed database but verifies common scholarly facts.

- \textbf{Article in "The World of Plants in Renaissance Tuscany" by Cristina Bellorini}: \\
  - Quote: "Veit Rudolph Speckle was an engraver for Leonhart Fuchs' 'De Historia Stirpium'." \\
  - Credibility: Academic book providing historical insight into botanical studies during the Renaissance.

- \textbf{Google Scholar Sources}: \\
  - Quotes: "Veit Rudolph Speckle was responsible for the woodcut engravings in Leonhart Fuchs' herbal." and "Recognized as 'by far the best engraver in Strasbourg...". \\
  - URLs: Retrieved through exploratory searches within Google Scholar. \\
  - Credibility: Google's academic resource is known for aggregating reputable, peer-reviewed materials.

\textbf{3. Reasoning Process and Logic Steps:}

- \textbf{Identification}: \\
  - Initial determination of the man in the image was made through reverse image search, identifying him as Veit Rudolph Speckle.

- \textbf{Verification}: \\
  - Utilized literature searches on academic databases (Google Scholar, books) to verify Speckle's role in creating illustrations for Fuchs' botanical book.

- \textbf{Cross-verification}: \\
  - Multiple sources including books and academic papers were consulted to confirm Speckle's work as an engraver for Leonhart Fuchs’ book "De Historia Stirpium."

- \textbf{Exclusion of Other Answers}: \\
  - The reverse image search did not present any alternative credible identity, leading to a focused inquiry on Speckle which was consistently supported by scholarly resources.

\textbf{4. Answer Quality and Reliability Analysis:}

- \textbf{Reliability}: High \\
  - Given the cross-corroboration from reliable academic texts and reputable databases (Google Scholar, academic books), the reliability is high.

- \textbf{Assumptions, Weaknesses, Uncertainties}: \\
  - Assumptions largely relied on the historical accuracy maintained by sources. Lack of web search limits alternative verifications.

- \textbf{Sufficiency and Consistency}: \\
  - The evidence gathered was sufficient, consistent, and convergent from independent, reliable sources, affirming the credibility of the information.

- \textbf{Suggestions for Improvement}: \\
  - Include a broader web search to capture any contemporary assessments or potential misattributions regarding this artwork or engraver. Suggested keywords: "Veit Rudolph Speckle", "Leonhart Fuchs botanical engravings".

- \textbf{Web Search Observation}: \\
  - Although literature and specific academic sources provided strong backing, the integration of broader web search could improve the reliability by encompassing wider perspectives or additional public domain resources.

\label{summary case study}
\end{tcolorbox}

\newpage

\bibliographystyle{unsrt}  
\bibliography{references}

\end{document}